\documentclass[letterpaper, 10pt, conference]{ieeeconf}  
\IEEEoverridecommandlockouts                              
\usepackage{color,xcolor}
\usepackage{epsfig}
\usepackage{graphicx}

\usepackage{adjustbox}
\usepackage{array}
\usepackage{booktabs}
\usepackage{colortbl}
\usepackage{float,wrapfig}
\usepackage{hhline}
\usepackage{multirow}
\usepackage{subcaption} 
\usepackage[font={small}]{caption}

\usepackage{amsmath,amsfonts,amssymb}
\usepackage{bm}
\usepackage{nicefrac}
\usepackage{microtype}

\usepackage{changepage}
\usepackage{extramarks}
\usepackage{fancyhdr}
\usepackage{lastpage}
\usepackage{setspace}
\usepackage{soul}
\usepackage{xspace}

\usepackage{url}

\usepackage{algorithm, algorithmic}
\usepackage{enumerate}
\usepackage{todonotes} 
\usepackage{gensymb}

\newcolumntype{L}[1]{>{\raggedright\let\newline\\\arraybackslash\hspace{0pt}}m{#1}}
\newcolumntype{C}[1]{>{\centering\let\newline\\\arraybackslash\hspace{0pt}}m{#1}}
\newcolumntype{R}[1]{>{\raggedleft\let\newline\\\arraybackslash\hspace{0pt}}m{#1}}

\newcommand{\sect}[1]{Sec.~\ref{#1}}

\newcommand{\eqn}[1]{Eqn.~\ref{#1}}

\newcommand{\fig}[1]{Fig.~\ref{#1}}
\newcommand{\figs}[1]{Figs.~\ref{#1}}
\newcommand{\tbl}[1]{Table~\ref{#1}}

\newcommand{\ignorethis}[1]{}
\newcommand{\norm}[1]{\lVert#1\rVert}

\DeclareMathOperator*{\argmin}{arg\,min}

\makeatletter
\DeclareRobustCommand\onedot{\futurelet\@let@token\@onedot}
\def\@onedot{\ifx\@let@token.\else.\null\fi\xspace}

\def\eg{\emph{e.g}\onedot} 
\def\ie{\emph{i.e}\onedot} 
 
 \def\vs{\emph{vs}\onedot}
 
\def\etal{\emph{et al}\onedot}
\makeatother

\definecolor{MyDarkBlue}{rgb}{0,0.08,1}
\definecolor{MyDarkGreen}{rgb}{0.02,0.6,0.02}
\definecolor{MyDarkRed}{rgb}{0.8,0.02,0.02}
\definecolor{MyDarkOrange}{rgb}{0.40,0.2,0.02}
\definecolor{MyPurple}{RGB}{111,0,255}
\definecolor{MyRed}{rgb}{1.0,0.0,0.0}
\definecolor{MyGold}{rgb}{0.75,0.6,0.12}
\definecolor{MyDarkgray}{rgb}{0.66, 0.66, 0.66}

\title{\LARGE \bf
Augmenting Physical Simulators with Stochastic Neural Networks:\\ Case Study of Planar Pushing and Bouncing}

\author{Anurag Ajay$^{1}$, Jiajun Wu$^{1}$, Nima Fazeli$^{2}$, Maria Bauza$^{2}$, \\ Leslie P. Kaelbling$^{1}$, Joshua B. Tenenbaum$^{1}$, Alberto Rodriguez$^{2}$
\thanks{$^{1}$ A. Ajay, J. Wu, L. P. Kaelbling, and J. B. Tenenbaum are with the Computer Science and Artificial Intelligence Laboratory at Massachusetts Institute of Technology, Cambridge, MA, USA}%
\thanks{$^{2}$ N. Fazeli, M. Bauza, and A. Rodriguez are with the Department of Mechanical Engineering at Massachusetts Institute of Technology, Cambridge, MA, USA}%
}

\begin{document}

\maketitle
\thispagestyle{empty}
\pagestyle{empty}

\begin{abstract}

An efficient, generalizable physical simulator with universal uncertainty estimates has wide applications in robot state estimation, planning, and control. In this paper, we build such a simulator for two scenarios, planar pushing and ball bouncing, by augmenting an analytical rigid-body simulator with a neural network that learns to model uncertainty as residuals. Combining symbolic, deterministic simulators with learnable, stochastic neural nets provides us with expressiveness, efficiency, and generalizability simultaneously. Our model outperforms both purely analytical and purely learned simulators consistently on real, standard benchmarks. Compared with methods that model uncertainty using Gaussian processes, our model runs much faster, generalizes better to new object shapes, and is able to characterize the complex distribution of object trajectories.

\end{abstract}

\section{Introduction}
\label{sec:intro}

Simulators are an essential for the development of robot systems. An important class of systems in robotics is well approximated by rigid-body dynamics; relatively mature, fast, and general-purpose simulators have been developed for these systems. These simulators (\eg, ODE~\cite{Smith2006Open}, Bullet~\cite{Coumans2010Bullet}, MuJoco~\cite{Todorov2012MuJoCo}) rely on approximate and efficient dynamics models and do not reason about uncertainty explicitly. 

These simulators are an important tool, yet their practicality has been limited due to discrepancies between their predictions and real-world observations. A major source of mismatches is the contact models used in these simulators. Contact is a complex physical interaction with near impulsive forces over a small duration of time that involves local deformations and vibrations. Matters are complicated by the sensitivity of contact outcomes to initial conditions. These models are coarse approximations to contact, and recent studies (\cite{Kolbert2016}, \cite{Yu2016More}, \cite{Fazeli2017Fundamental}) have shown the discrepancies between their predictions and real-world data. Further, Fazeli~\etal\cite{Fazeli2017Fundamental} showed that there exist real-world contact outcomes which the models are unable to predict for any choice of their parameters. This suggests that generating uncertainty by defining distributions over contact parameters does not yield a sufficiently rich and descriptive distribution over outcomes.

\begin{figure}[t]
    \centering
    \includegraphics[angle=270,origin=c,width=0.85\linewidth]{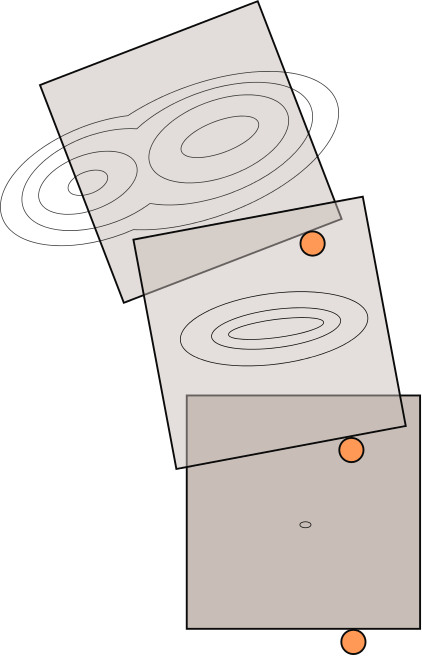}
    \vspace{-35pt}
    \caption{The motion of an object being pushed appears stochastic and possibly multi-modal due to imperfections in contact surfaces, non-uniform coefficient of friction, stick/slip transitions, and micro surface interactions. In this study, we propose to augment analytical models to more accurately predict such outcomes while reasoning about uncertainty.
    }
    \vspace{-20pt}
    \label{fig:teaser}
\end{figure}

In this study, we provide a framework for augmenting analytical motion models with empirical data that has higher accuracy while capturing uncertainty in predictions. We achieve this by using a novel type of recurrent neural networks, namely decoupled conditional variational recurrent neural nets, to learn the residual errors made by the analytical models. Once the neural networks are trained, they can correct model predictions and provide distributions over possible outcomes. 

We demonstrate the efficacy of the data-augmented stochastic simulation framework in two cases: 1) a toy bouncing ball problem, and 2) planar pushing with a single point pusher using the empirical dataset from Yu~\etal\cite{Yu2016More}. First, we use the toy problem to illustrate the implementation and details of the proposed model in simulation. We then move to the experimental planar pushing dataset to demonstrate the ability of the approaches to capture real-world data. We show that the data-augmented model outperforms its purely analytical and purely data-driven counterparts. Further, we demonstrate that this approach is data-efficient, as learning residuals is an easier and better formulated problem than learning full motion models. Experiments also suggest that the learned residual model generalizes better to different shapes than the pure learning-based dynamics model. The data-augmented residual model can reason about uncertainty, which plays an important role in planning and control.

\section{Related Work}
\label{sec:related}

\subsection{Models for Planar Pushing}

Planar pushing is an important instance of planar manipulation, in which the robot moves objects on a horizontal surface through a set of pushes, in particular for objects that are too heavy or too large to be picked up by the robot. To model planar pushing,  Goyal~\etal~\cite{Goyal1991} proposed the notion of ``Limit Surfaces'' (LS) as an invertible mapping between a push and a consistent set of friction forces and object motions. Given the object's current pose and force applied to it, the LS predicts the subsequent motion assuming quasi-static motion. The LS assumes a pressure distribution over the contact patch and Coloumb friction law; it integrates over all possible instantaneous centers of rotation for the object to yield the mapping. 

In general, the LS does not have a closed form solution; however, Howe and Cutkosky~\cite{howe1996practical} showed that the LS can be approximated by an ellipsoid for uniform pressure distributions, constant friction coefficient, and quasi-static motions. Lynch~\etal~\cite{lynch1992manipulation} used the ellipsoidal LS to develop a motion model to predict object motion given pusher motion. The ellipsoidal LS has been used for planar push control by Hogan and Rodriguez~\cite{Hogan2016} and for shape reconstruction by Yu~\etal~\cite{Yu2015Shape}. In this study we also use the model proposed by Lynch~\etal \cite{lynch1992manipulation} as our analytical motion model. 

\subsection{Learning Contact Dynamics}

Recently, researchers have looked towards data-driven techniques to complement existing analytical models and/or learn dynamics directly from data. The work by Kloss~\etal~\cite{kloss2017combining} is the closest to ours, where the authors trained a neural network that provides input to an analytical model. In this framework, the output of the analytical model is used as the prediction; the neural network learns the best input parameters to maximize the performance of the analytical model. A benefit of this approach is that the model predictions are always feasible because of the analytical model, but the approach is deterministic, and relies on the expressiveness of the analytical model. 

In the planar pushing case, these models may be sufficiently expressive to span the full range of outcomes, but this is not always the case in other contact interactions as shown by Fazeli~\etal~\cite{Fazeli2017Fundamental}. Further, the models in \cite{kloss2017combining} only make single-step predictions---an approach that may not work well for long-horizon predictions due to compounding errors at each time step. In our paper, we use the analytical model as an approximation to the push outcomes, and learn a residual model that makes corrections to its output. We are thus not limited by the model's expressivity, as the neural network can make corrections outside the predictive range of the models. Further, we learn a stochastic recurrent network that makes long-horizon predictions in the form of a distribution over possible outcomes. We believe reasoning about the degree of confidence in outcome prediction can be used effectively in planning and control.

Fazeli~\etal~\cite{Fazeli2017Learning} also proposed to learn a residual model for prediction of empirical planar impacts. The residual learner in their paper is a Gaussian process and achieves significant improvement over the analytical contact models in terms of its prediction accuracy. Gaussian processes are however limited to Gaussian predictive distributions and are computationally slower, compared with neural networks. Further, the authors also did not study the effect of making long-term predictions, as their focus is on individual impact prediction accuracy. Zhou~\etal~\cite{zhou2016convex} supplied a data-efficient approach to model the frictional interaction between an object and a support surface, by directly approximating the mapping between frictional wrench and slipping twist. Later, Zhou~\etal~\cite{zhou2017fast} extended the model to simulate parametric variability in planar pushing and grasping.

Byravan and Fox~\cite{byravan2017se3} showed how to design a neural network to predict rigid-body motions in a planar pushing scenario. In this study, as a robot pushes an object, the neural network differentiates between the object and the table. The neural network makes predictions by explicitly predicting $SE(3)$ transformations and jointly learning the full motion model and the observation model. This approach is still deterministic and does not use any more physics knowledge.

\vspace{-0.07cm}
\subsection{Uncertainty Modeling}

Reasoning about the uncertainty in actions and motions is a powerful tool in planning and control~\cite{bauza2017probabilistic,babaeizadeh2017stochastic,Xue2016Visual, bauza2017gp}. In the context of planar manipulation, Bauza and Rodriguez~\cite{bauza2017gp} used Gaussian processes to learn the motion model of planar shapes and to propagate uncertainty using the GP-SUM algorithm. The GP-SUM algorithm is a hybrid Bayes and particle filter; it exploits the Gaussian structure of the motion model to efficiently approximate the distribution over outcomes as a mixture of Gaussians. Bauza and Rodriguez~\cite{bauza2017gp} showed that pushing can exhibit multi-modality and their approach is able to capture it. We use the model and algorithm from~\cite{bauza2017gp} as benchmarks for our approach and compare the two on the MIT push dataset~\cite{Yu2016More}.

A practical example of using the knowledge of uncertainty in planar manipulation was introduced by Zhou~\etal~\cite{zhou2017probabilistic}. They proposed a probabilistic algorithm that generates sequential actions to iteratively reduce uncertainty of objects in the plane, before grasping it with a parallel jaw gripper.
\section{Formulation}
\label{sec:formulation}

\begin{figure*}[t]
    \centering
    \includegraphics[width=\linewidth]{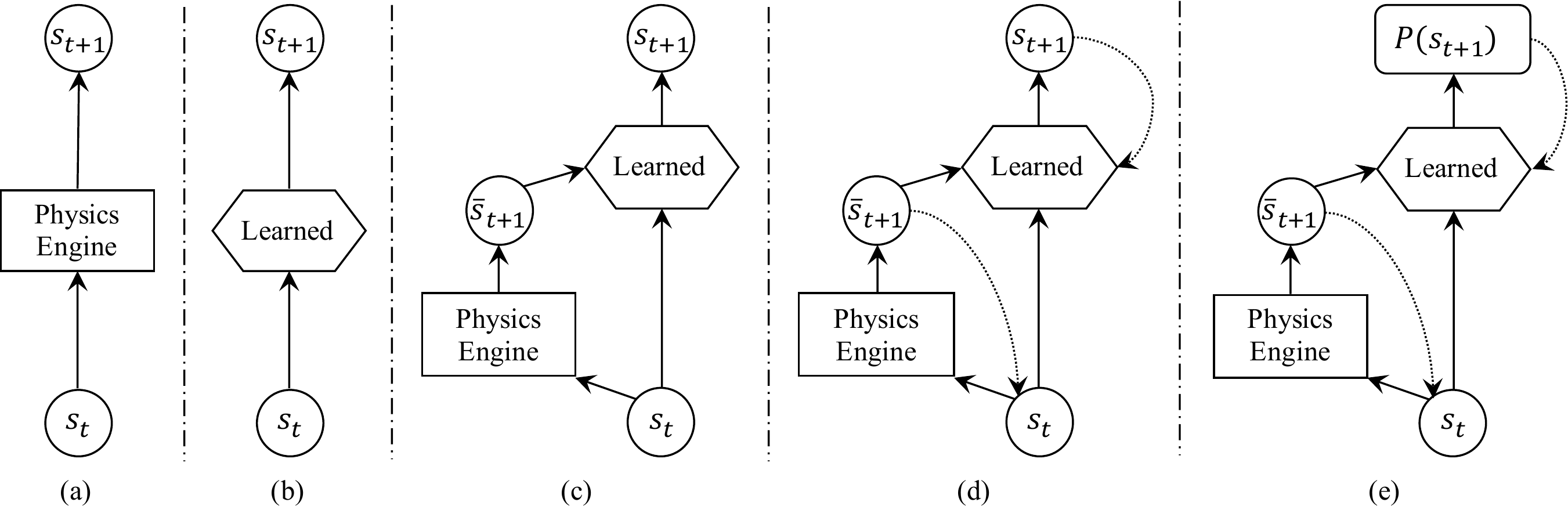}
    \vspace{-18pt}
    \caption{Model classes: (a) physics-based analytical models; (b) data-driven models; (c) data-augmented residual models; (d) recurrent data-augmented residual models; and (e) stochastic recurrent data-augmented residual models.}
    \vspace{-18pt}
    \label{fig:model}
\end{figure*}

In this section we provide the details of our proposed data-augmented stochastic simulation framework. The simulation framework has two components: an analytical model and a data-driven residual model. We first define each component; we then provide a detailed exposition of the data-driven residual model and its role as a method to improve simulation accuracy and to maintain a belief over states.

Let $S$ represent the state space, $A$ represent the action space, and $(s,a,s')$ represent a state-action-state tuple, where $s, s' \in S$, $a \in A$, and $s'$ is the state obtained after applying action $a$ in state $s$. A dynamics model is a function $f: S \times A \rightarrow S$ that predicts the next state given the current action and state:
$f(s, a) \approx s', \;\; s, s' \in S, \;\; a \in A$.

We distinguish between two classes of dynamic models: physics-based analytical models and data-driven models.

\vspace{-0.10cm}
\subsection{Physics-Based Analytical Models}
\vspace{-0.05cm}

These models are constructed from the laws of physics, domain knowledge, and convenient approximations often made for mathematical tractability. In this paper, we also refer to them as Physics Engines and use the terms interchangeably, though in practice physics engines may not be faithfully implementing the mathematical models. Generally, these models work well close to their assumptions and in structured environments, but their performance degrades as we move away from their nominal working conditions. Further, finding tractable models for complex tasks is difficult and requires extensive domain specific expertise. For the rest of this paper, let $f_p: S \times A \rightarrow S$ represent the analytical model.

\vspace{-0.10cm}
\subsection{Data-Driven Models}
\vspace{-0.05cm}

Rather than being hand engineered, these models are learned using data collected from the real world. They can be either parametric (\eg, neural networks) or non-parametric (\eg, Gaussian processes). For the purpose of discussion, let's assume a parametric model represented by $f_{\theta}: S \times A \rightarrow S$, where $\theta$ is the parameter vector. The model is learned using data collected from the real world; for example, the robot may take actions according to a fixed pushing policy and collect $(s,a,s')$ tuples that represent the states of the object being pushed and the motion of the pusher. After collecting data $\{(s_t, a_t, s_{t+1})\}_{t=0}^{T-1}$, we solve the following optimization problem to obtain optimal parameters for the model:
\vspace{-5pt}
\begin{equation}
\theta^* = \argmin\limits_{\theta} \sum_{t=0}^{T-1} \|f_{\theta}(s_t, a_t) - s_{t+1}\|_2^2 + \lambda \|\theta\|_2^2
\vspace{-5pt}
\end{equation}
where $\lambda$ is a constant for regularization. After obtaining $\theta^*$, we use $f_{\theta^*}$ as the representation of our motion model. While this approach requires no hand-engineering and directly learns from the data without making any assumptions, it does not make use of any domain knowledge, and consequently may require many examples to learn.

\vspace{-0.10cm}
\subsection{Data-Augmented Residual Models}
\vspace{-0.05cm}

We leverage advantages of both model classes and develop a new hybrid class of models, which we call \emph{data-augmented residual models}, by combining a physics engine with a data-driven model. In this modeling framework, the data-driven part of the model takes the current state-action pairs and the predictions made by the physics engine as input, and effectively learns the discrepancy between analytical model predictions and real-world data (\ie the residual). If $f_r$ represents the data-augmented residual model, $f_p$ represents its physics engine, and $f_{\theta}$ represents its residual component, we have
$f_r(s, a) = f_{\theta}(f_p(s,a), s, a) \approx s'$.
Intuitively, the residual model refines the physics engine's guess using the current state and action.

\vspace{-0.10cm}
\subsection{Recurrent Data-Augmented Residual Models}
\vspace{-0.05cm}

Planning and control require long-horizon predictions of future states of the world, given actions taken by an agent using dynamic models. No matter how accurate the model is, it will have some error which will compound over a sequence of time steps. Moreover, the data-driven and data-augmented models are trained using data from real world trajectories. While simulating the future, these dynamics models will recursively use their own prediction as input for the next time step. As there will be error in their predictions at each time step, the input data given during simulation phase will have a different distribution than the input data during the training phase. This creates data distribution mismatch between training and test (or simulation) phases for both data-augmented residual models and purely data-driven models. 

To address this problem, we propose to use a recurrent data-augmented residual model, trained to predict the entire trajectory based on an initial state and an action sequence. The \emph{recurrent data-augmented residual model} consists of two components: a physics engine and a recurrent data-augmented residual model. The physics engine takes in the initial state and a sequence of actions at every time step; it generates an entire trajectory which serves as a good initial guess for the recurrent residual model. The residual model takes the initial state, a sequence of actions, and the trajectory predicted by the physics engine; it then predicts the next state. If $f^R_r$ represents the data-augmented recurrent residual model, $f_p$ represents its physics engine, and $f^R_{\theta}$ represents its residual component, we have
\begin{gather}
f^R_r(\bar{s}_t, \hat{s}_t, a_t) = f^R_{\theta}(f_p(\bar{s}_t, a_t), \hat{s}_t, a_t) = \hat{s}_{t+1} \approx s_{t+1},\\
f_p(\bar{s}_t, a_t) = \bar{s}_{t+1}, \quad  \bar{s}_0 = \hat{s}_0 = s_0,
\end{gather}
where ${(\hat{s}_t)_{t=0}^{T-1}}$ is the predicted trajectory. The model is fully differentiable and can be trained by minimizing $\min_{\theta} \sum_{t=0}^{T-1} \|\hat{s}_t - s_t\|_2^2 + \lambda \|\theta\|_2^2$.

\vspace{-0.10cm}
\subsection{Stochastic Recurrent Data-Augmented Residual Models}
\vspace{-0.05cm}

No model is perfect, therefore the ability to provide a measure of uncertainty over possible future states is an important capability, allowing better informed planning and control. To this end, we formulate a stochastic model. Let $f^R_r$ represents stochastic data-augmented recurrent residual model, we have
\begin{gather}
f^R_r(\bar{s}_t, \hat{s}_t, a_t) = f^R_{\theta}(f_p(\bar{s}_t, a_t), \hat{s}_t, a_t) = \hat P_{{s}_{t+1}}(\cdot), \label{eqn:distr} \\
f_p(\bar{s}_t, a_t) = \bar{s}_{t+1}, \quad \bar{s}_0 = \hat{s}_0 = s_0,
\end{gather}
where $\{\hat{P}_{s_t}(\cdot)\}_{t=0}^{T-1}$ is the predicted trajectory distribution. The model is also differentiable and can be trained by minimizing
$\min_{\theta} \sum_{t=0}^{T-1} \log\hat{P}_{{s}_t}(s_t | \theta) + \lambda \|\theta\|_2^2$.

\section{Methods}
\label{sec:method}

We now present how we realize the stochastic recurrent data-augmented residual model by providing an overview of the components and the way they are connected. 

\vspace{-0.1cm}
\subsection{The Analytical Push Motion Model}
\label{sec:ana}

As mentioned in \sect{sec:related}, we use the model proposed in \cite{lynch1992manipulation} as our analytical pushing motion model (physics engine). The motion model takes the current configuration of the object and the pusher motion, and returns the predicted object motion at each time step. To make predictions, the motion model first computes a motion cone for the current pusher velocity and object configuration. Next, depending on whether the pusher velocity lies inside or outside of the cone, the model identifies the contact as sticking or sliding respectively. Finally the model computes the object motion using the sticking/sliding classification and the ellipsoidal LS.

In the experimental setup of the planar push dataset \cite{Yu2016More}, the time history of object and robot pusher motion are recorded. We can compute the analytical model predictions using the current configuration of the object and robot pusher motion. In doing this, we observe discrepancies between the model prediction and the measured data. The discrepancy is due in part to the ellipsoidal approximation, and in part to variations in coefficients of friction across the surface, micro-interactions between the object and surface, and potentially anisotropic frictional properties. The latter effects are impractical to model analytically and difficult to predict ahead of time.

\vspace{-0.1cm}
\subsection{Stochastic Neural Networks}

\begin{figure*}[t]
    \centering
    \includegraphics[width=\linewidth]{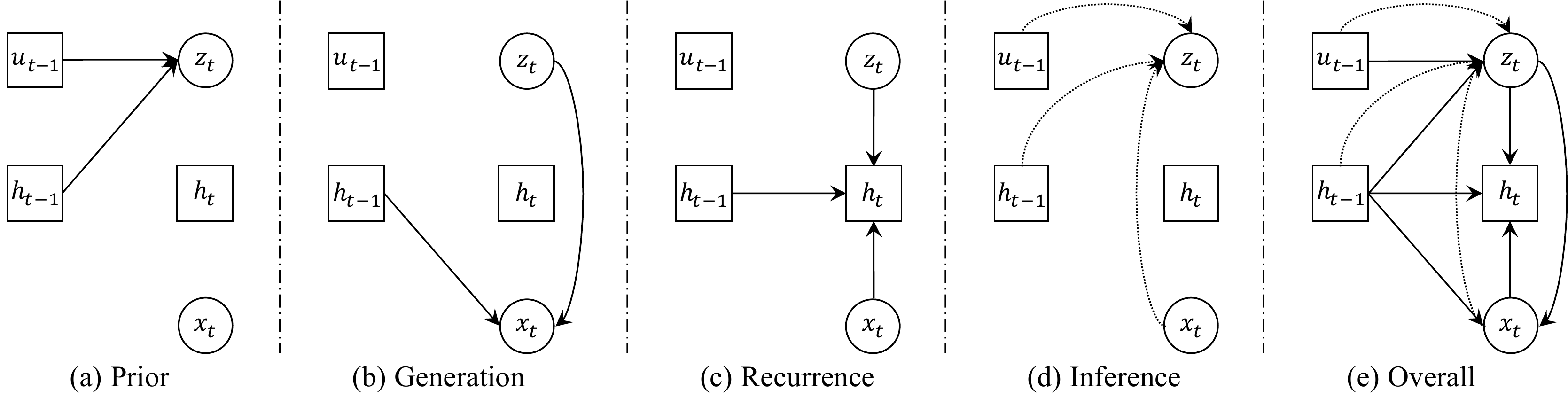}
    \vspace{-18pt}
    \caption{Illustrations for conditional VRNNs: (a) the prior model of the latent representation $z_t$; (b) the generation model for $x_t$; (c) the recurrence model for $h_t$; (d) the inference model for $z_t$; and (e) the overall conditional VRNN design.}
    \vspace{-18pt}
    \label{fig:network}
\end{figure*}

We implement our recurrent data-augmented residual model as a GRU~\cite{Cho2014GRU}, a widely used recurrent network for modeling long-term correlations. The model is however deterministic. The simplest way to incorporate stochasticity is to model $P_{{s}_t}(.)$ in \eqn{eqn:distr} as a Gaussian distribution, \ie, $\hat P_{{s}_t}(.) = \mathcal{N}(\hat\mu_{s_t}, \hat\sigma_{s_t})$. However, this limits our model's ability to characterize complex distributions in real world. 

Chung~\etal~\cite{chung2015recurrent} proposed to incorporate variational autoencoders~\cite{Kingma2014Auto} into recurrent nets and named their model variational RNNs (VRNNs). A VRNN supports modeling highly complex distributions over time. Their model however cannot be conditioned on additional inputs such as control variables (\eg push forces). We instead embed a conditional variational autoencoder into our GRU. It therefore becomes a variant of VRNNs, namely \emph{Conditional VRNNs}. 

\vspace{-0.1cm}
\subsubsection{Variational Recurrent Neural Networks}

VRNNs are recurrent generative models used for modeling multi-modal trajectories. It has three interconnected components: priors, an encoder, and a decoder. Suppose we represent a given trajectory as $\{x_t\}_{t=0}^T$. During training, the encoder takes the trajectory as input and infers latent random variables $\{z_t\}$ as
\begin{align}
\label{eqn:post}
    P(z_t | x_t) &\sim \mathcal{N}(\mu_{z,t}, \sigma_{z,t}), \\ 
    [\mu_{z,t}, \sigma_{z,t}] &= \varphi^{\text{enc}}(\varphi^x(x_t), h_{t-1}),
\end{align}
where $\varphi^{\text{enc}}$ represents the encoder, $\varphi^x$ is a function that extracts features of $x_t$, and $h$ is the hidden vector in the GRU. We then sample the latent random variable from the above distribution using a reparameterization trick~\cite{Kingma2014Auto}, formulated as
$z_t = \mu_{z,t} + \epsilon_t\times\sigma_{z,t}, \;\; \epsilon_t \sim \mathcal{N}(0,I)$.
After that, the decoder uses the sampled latent variable $z_t$ to reconstruct the trajectory, following
$P(x_t | z_t ) \sim \mathcal{N}(\mu_{x,t}, \sigma_{x,t})$,
where
\begin{equation}
\label{eqn:decode}
[\mu_{x,t}, \sigma_{x,t}] = \varphi^{\text{dec}}(\varphi^z(z_t), h_{t-1}).
\end{equation}
Here, $\varphi^{\text{dec}}$ is the decoder and $\varphi^z$ is a feature extractor for $z_t$. 

In a VAE, we enforce the distribution of the latent vector $\{z_t\}$ to be close to a prior distribution~\cite{Kingma2014Auto}. In VRNN, the prior $\varphi^{\text{prior}}$ is learned and follows the distribution
\begin{equation}
\label{eqn:prior}
P(z_t) \sim \mathcal{N}(\mu_{0,t}, \sigma_{0,t}), \quad \text{where} \;\; [\mu_{0,t}, \sigma_{0,t}] = \varphi^{\text{prior}}(h_{t-1}).
\end{equation}
Finally, the RNN $f^{\text{RNN}}$ updates its state as
\begin{equation}
\label{eqn:state}
h_t = f^{\text{RNN}}(\varphi^x(x_t), \varphi^z(z_t), h_{t-1}).
\end{equation}
VRNN is trained by minimizing
\begin{align}
\label{eqn:loss}
\sum_{t=1}^T & D_{\text{KL}}\left(\mathcal{N}(\mu_{z,t}, \sigma_{z,t} || \mathcal{N}(\mu_{0,t}, \sigma_{0,t})\right) \nonumber\\
& - \frac{1}{L} \sum_{t=1}^T \sum_{i=1}^L P(x_t | \mu_{z,t} + \epsilon_{i,t}\times\sigma_{z,t}) + \lambda \|\theta\|_2^2,
\end{align}
where $\epsilon_{i,t} \sim \mathcal{N}(0,I)$, $\lambda$ is a regularization constant, and $\theta$ is a vector containing all the parameters in our model. Once the VRNN is trained, we use the prior to sample latent random variables and use them to generate trajectories.

\subsubsection{Conditional VRNNs}

We want a VRNN to be conditioned on a sequence $\{u_t\}_{t=0}^T$ (\eg, the control inputs). To this end, the posterior distribution of $z_t$ (\eqn{eqn:post}) is now
$P(z_t | x_t, u_t ) \sim \mathcal{N}(\mu_{z,t}, \sigma_{z,t})$,
where 
\begin{equation}
[\mu_{z,t}, \sigma_{z,t}] = \varphi^{\text{enc}}(\varphi^x(x_t), \varphi^u(u_t), h_{t-1}).
\end{equation}

The prior distribution (\eqn{eqn:prior}) also becomes conditional
$P(z_t | u_t) \sim \mathcal{N}(\mu_{0,t}, \sigma_{0,t})$,
where 
\begin{equation}
[\mu_{0,t}, \sigma_{0,t}] = \varphi^{\text{prior}}(\varphi^u(u_t), h_{t-1})
\end{equation}
And the state update equation (\eqn{eqn:state}) becomes
\begin{equation}
h_t = f^{\text{RNN}}(\varphi^x(x_t), \varphi^z(z_t), \varphi^u(u_t), h_{t-1}).
\end{equation}

Here, the decoder (\eqn{eqn:decode}) does not depend on $\{u_t\}_{t=0}^T$ because the latent vectors $\{z_t\}_{t=0}^T$ already have the capacity to contain all information about the control sequence $\{u_t\}_{t=0}^T$.

\subsubsection{Decoupled Conditional VRNNs}

In our experiments, we predict trajectories whose length varies from 100 to 1000. Because training and evaluating a VRNN becomes slower with these long trajectories, we propose an approximation to conditional VRNNs, which we call \emph{Decoupled Conditional VRNNs}. A conditional VRNN is slow, because updates in a RNN have temporal dependence. However, we observe that the encoding, decoding, and prior networks are not inter-dependent: for example, the encoder only needs $\epsilon_t$ to sample $z_t$ internally; it does not take signals from the decoding and the prior networks. Thus, in DCVRNNs, we disentangle the model into three recurrent neural nets, one each for priors, the encoder, and the decoder. Specifically, we first have the Gaussian noises sampled as $\epsilon_t \sim \mathcal{N}(0,I)$. Then the equation for the encoder becomes
$P(z_t | x_t, u_t ) \sim \mathcal{N}(\mu_{z,t}, \sigma_{z,t})$,
where
\begin{equation}
h_t, [\mu_{z,t}, \sigma_{z,t}] = f^{\text{RNN}}_{\text{enc}}(\varphi^x(x_t), \varphi^u(u_t), \epsilon_t, h_{t-1}).
\end{equation}
The decoder is now
$P(x_t | z_t ) \sim \mathcal{N}(\mu_{x,t}, \sigma_{x,t})$,
where
\begin{equation}
h_t, [\mu_{x,t}, \sigma_{x,t}] = f^{\text{RNN}}_{\text{dec}}(\varphi^z(z_t), h_{t-1}).
\end{equation}
The prior is now
$P(z_t | u_t) \sim \mathcal{N}(\mu_{0,t}, \sigma_{0,t})$,
where
\begin{equation}
 h_t, [\mu_{0,t}, \sigma_{0,t}] = f^{\text{RNN}}_{\text{prior}}(\varphi^u(u_t), \epsilon_t, h_{t-1}).
\end{equation}
The loss function remains unchanged.

We use a DCVRNN as our stochastic data-augmented residual model by having
\begin{equation}
x_t = s_t, \quad u_t = [s_0, a_t, \hat{s_{t+1}}], \quad \hat{s}_{t+1} = f_p(\hat{s}_t, a_t),
\end{equation}
where $s_t$ represents the state at time t, $a_t$ represents the action at time t, $f_p$ represents the physics engine, and $\hat{s}_t$ represents the state predicted by the physics engine.

\section{Experiments}
\label{sec:exp}

We study two scenarios: ball bouncing and planar pushing. We first generate synthetic data of ball bouncing and use them as an illustrative example to demonstrate the efficacy of our model. We then evaluate our model on the MIT Push dataset~\cite{Yu2016More} and compare it with baselines and state-of-the-art motion models. We further present analyses on how well our model generalizes across shapes and materials.

\vspace{-0.17cm}
\subsection{Experiments with Bouncing Balls}

\begin{figure}[t]
    \centering
    \includegraphics[width=\linewidth]{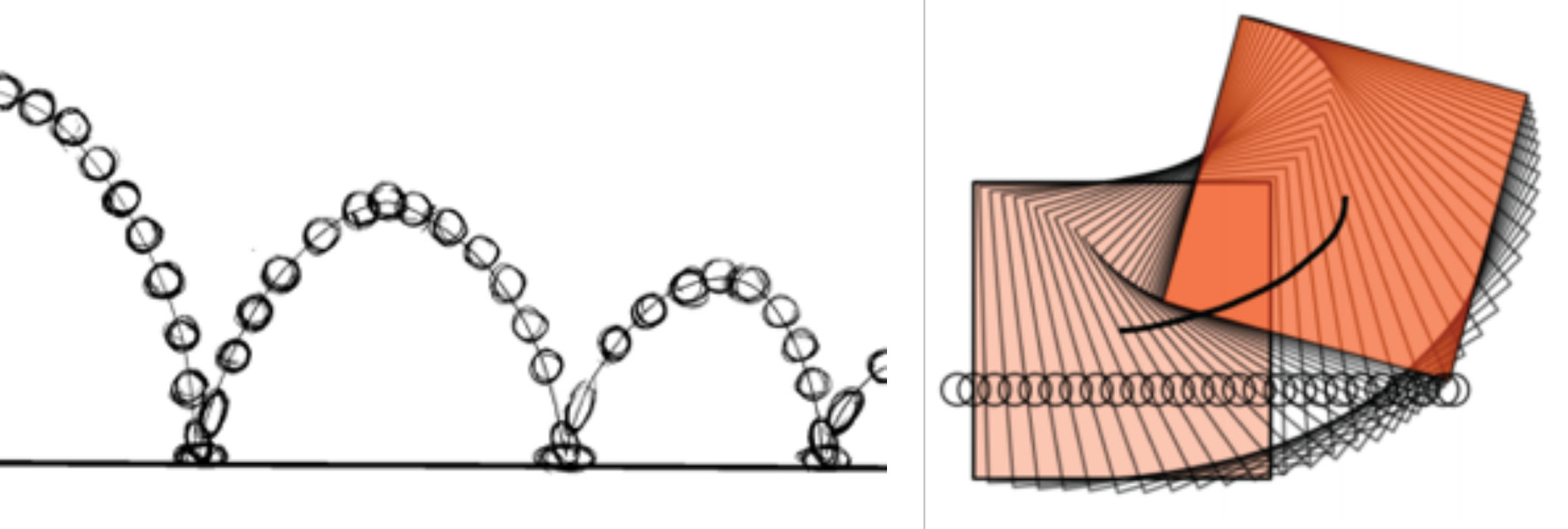}
    \vspace{-15pt}
    \caption{\small{The two scenarios: ball bouncing and planar pushing.}} 
    \vspace{-18pt}
    \label{fig:scenarios}
\end{figure}

\begin{table*}[t]
    \small
    \centering
    \begin{tabular}{lcccccccccc}
    \toprule
    & \multicolumn{4}{c}{Train} & \multicolumn{4}{c}{Test} \\
    \cmidrule(lr){2-5}\cmidrule(lr){6-9}
    Models & loss ($\times10^{-2}$) & trans (\%) & pos (mm) & rot (deg) & loss ($\times10^{-2}$) & trans (\%) &  pos (mm) & rot (deg)\\
    \midrule
    Zero & N/A & N/A & N/A & N/A & N/A & 99.99 & 359.44 & 49.46 \\
    Physics & N/A & N/A & N/A & N/A & N/A & 1.93 & 6.91 & 7.71 \\ 
    Neural & 0.41 & 0.72 & 2.42 & 1.85 & 0.68 & 0.84 & 2.81 & 2.48 \\
    Hybrid & 0.36 & 0.54 & 1.86 & 1.73 & 0.47 & 0.60 & 2.04 & 2.03 \\
    \bottomrule
    \end{tabular}
    \caption{Our hybrid model achieves the best performance in both position and rotation estimation for \emph{rect1}, compared with methods that rely on physics engines or neural nets alone. Here we show results on both training and test sets, as well as the optimization losses. These numbers suggest that our Hybrid model is overfitting to the training set less than the pure Neural model. As we focus on long-term prediction, we include the Zero baseline to show the scale and the challenging nature of the problem.} 
    \vspace{-18pt}
    \label{tbl:results_push}
\end{table*}

\noindent{\bf Data. }
When a ball bounces against the ground, it may reach different heights due to the irregularities in the ground surface which leads to different coefficients of restitution. Here we simulate the process using PyBullet. Specifically, we choose a height randomly from [4m, 5m] and a coefficient of restitution for ball-ground interaction from $\mathcal{N}(0.5, 0.1)$. We then drop a ball of radius 0.5m from the sampled height, and record its height and vertical velocity for 400 time steps with a sampling frequency of 60Hz. For our physics engine, we create another PyBullet environment but fix the coefficient of restitution to 0.65. We want our physics engine to produce trajectories that are different from our training data, but serve as a good initial guess. 

\noindent{\bf Metrics. }
We use three metrics for evaluation. The first two are the average error in object height reported as a percentage (trans) and in absolute values with meters as unit (pos). The third is the average error on the object's vertical velocity (vel) in metres per second.

\noindent{\bf Methods. }
We compare with three baselines. The first is just the average translation and rotation over the dataset. This is equal to the error of always predicting zero movement, and we therefore name it Zero. The second (Physics) is the full deterministic, analytical model described above. The third (Neural) is to use the stochastic neural network alone without the simulator. Our full model (Hybrid) combines the simulator's and the network's predictions. For the stochastic Hybrid and Neural models, we sample 10 trajectories for each input and take their mean as our prediction. 

\noindent{\bf Setup. }
We implement our network in PyTorch. We train our network using the loss function in \eqn{eqn:loss}. We use the ADAM optimizer~\cite{Kingma2015Adam} with a learning rate of $10^{-3}$, a decay of 0.5 every 2,500 iteration for a total of 10,000 iteration, and a batch size of 100. Our training set contains 800 trajectories, while our test set has 100. For this experiment, $\varphi^x$, $\varphi^u$, $\varphi^z$ are all identity functions. $f^{\text{RNN}}_{\text{enc}}$ and $f^{\text{RNN}}_{\text{prior}}$ are both GRUs with 2 hidden layers and a hidden size of 16, followed by a linear layer of hidden size 4 for mean, and another parallel linear layer of hidden size 4 with softplus activation for standard deviation. $f^{\text{RNN}}_{\text{dec}}$ is a GRU with 2 hidden layers and a hidden size of 16, followed by a linear layer of hidden size 2 for mean, and another parallel linear layer of hidden size 2 with softplus activation for standard deviation. The decoder's standard deviation is kept fixed to identity.

\noindent{\bf Results. }
\tbl{tbl:results_ball} suggests that our model is able to outperform the baselines on the synthetic dataset. The Physics baseline, designed to be deterministic, is not performing very well as expected. However, with the help of the physics engine, our Hybrid model achieves much better performance compared to the Neural model, which learns everything from scratch. The intuition is that learning from a good guess makes the learning problem significantly easier.

\vspace{-0.15cm}
\subsection{Experiments on Planar Pushing}

\begin{table}[t]
    \centering
    \small
    \begin{tabular}{lccc}
    \toprule
    Models & trans (\%) & pos (m) & velocity (m/s$^2$) \\
    \midrule
    Zero & 100.00 & 0.64 & 1.60 \\
    Physics & 27.41 & 0.16 & 1.06 \\
    Neural & 9.16 & 0.058 & 0.43 \\
    Hybrid & 2.42 & 0.016 & 0.14 \\
    \bottomrule
    \end{tabular}
    \caption{Our hybrid model achieves the best performance in both position and velocity estimation of the ball, compared with methods that rely on physics engines or neural nets alone.}
    \vspace{-18pt}
    \label{tbl:results_ball}
\end{table}

\noindent{\bf Data. }
We use the MIT Push dataset~\cite{Yu2016More} for the scenario of planar pushing, which contains object pose and force recordings from real robot experiments. For uncertainty modeling in particular, we use the straight-line push experiment which was repeated 2,000 times. In this experiment, the object shape is a rectangle, the contact location is half way in between the block's center and edge, the contact is made perpendicular to the edge, the speed of the pusher is set to 20 mm/s with no acceleration, and the total pusher displacement is 15cm.

Part of the uncertainty comes of measurement noise, which we want to minimize. The object in the experiments is instrumented with reflective markers and tracked with Vicon motion tracking system. Vicon, when correctly calibrated, has 1mm or better accuracy with a unimodal distribution of noise, well approximated by a Gaussian. Because of its high fidelity, our model can focus on learning the uncertain in dynamics.

\noindent{\bf Metrics and methods. }
We use three metrics for evaluation, following Kloss~\etal~\cite{kloss2017combining}. The first two are the average Euclidean distance between the predicted and the ground truth object reported as a percentage relative to the initial pose (trans) and as absolute values (pos) in millimeters. The third is the average error of object rotation (rot) in degree. We compare with the same baselines as in the ball experiment, except that the physics engine is now the analytical model described in \sect{sec:ana}.

\begin{figure*}[t]
    \centering
    \begin{subfigure}[b]{0.32\linewidth}
    \includegraphics[width=\linewidth]{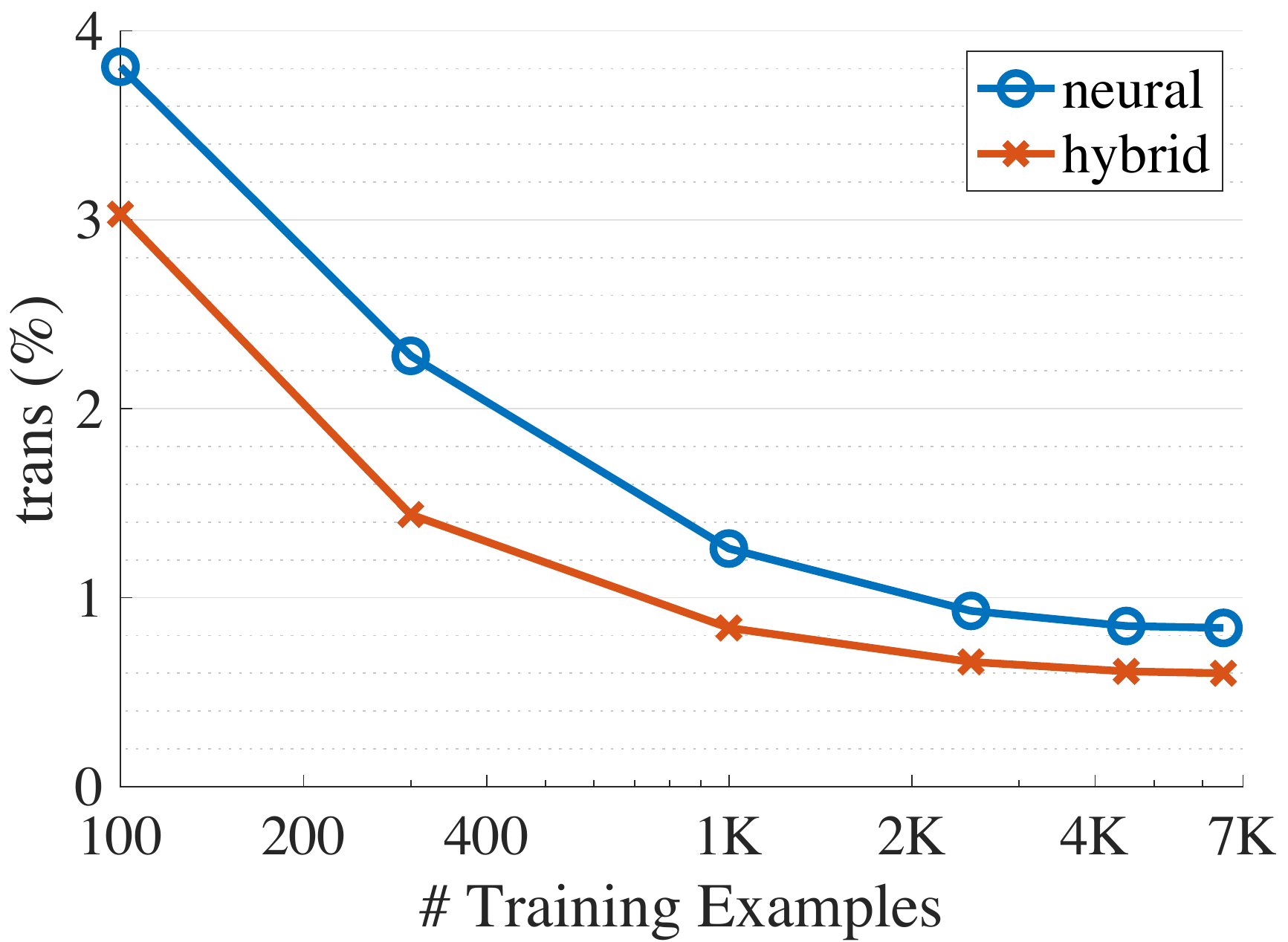}
    \caption{translation (\%)}
    \end{subfigure}
    \begin{subfigure}[b]{0.32\linewidth}
    \includegraphics[width=\linewidth]{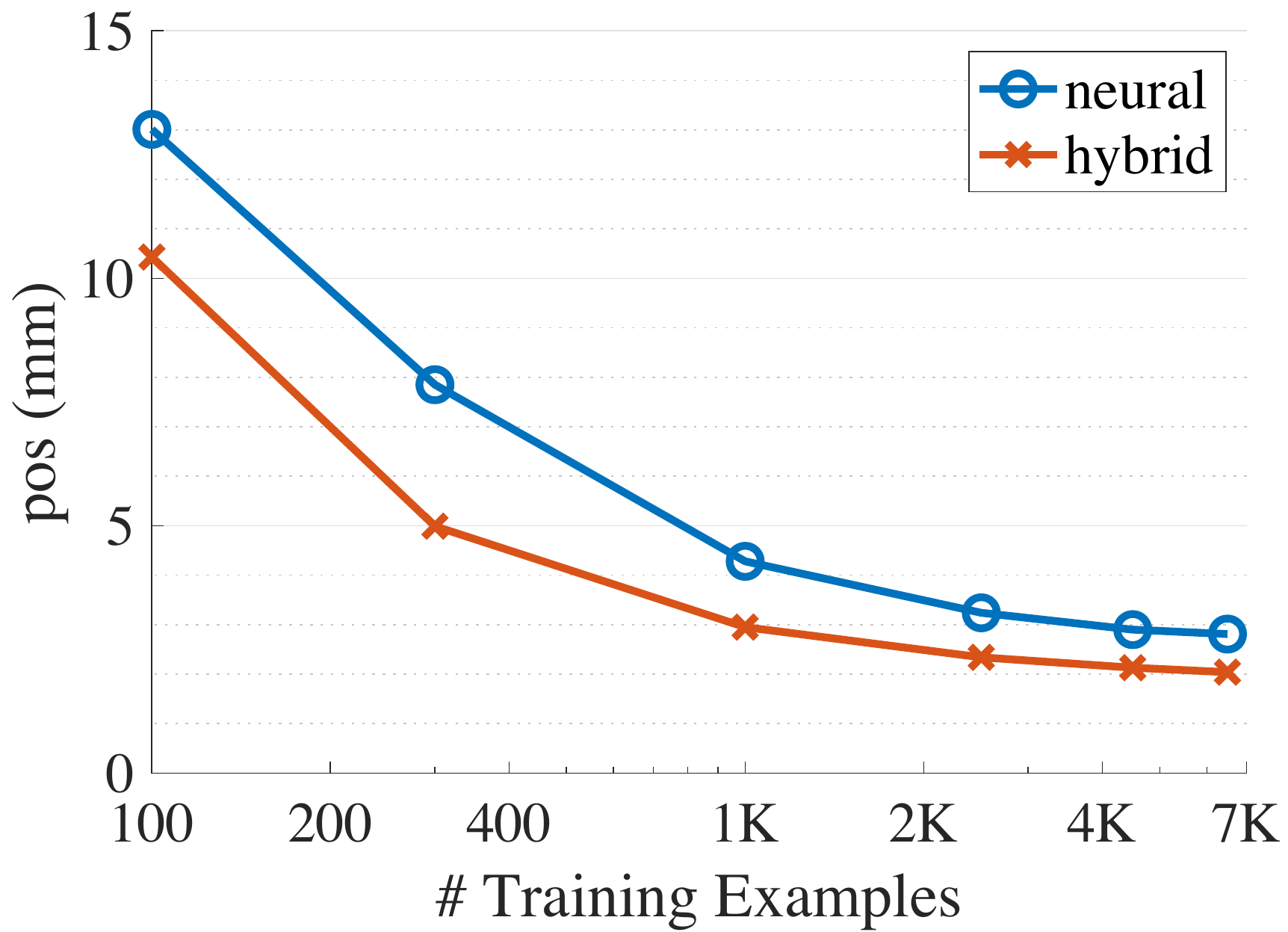}
    \caption{position (mm)}
    \end{subfigure}
    \begin{subfigure}[b]{0.32\linewidth}
    \includegraphics[width=\linewidth]{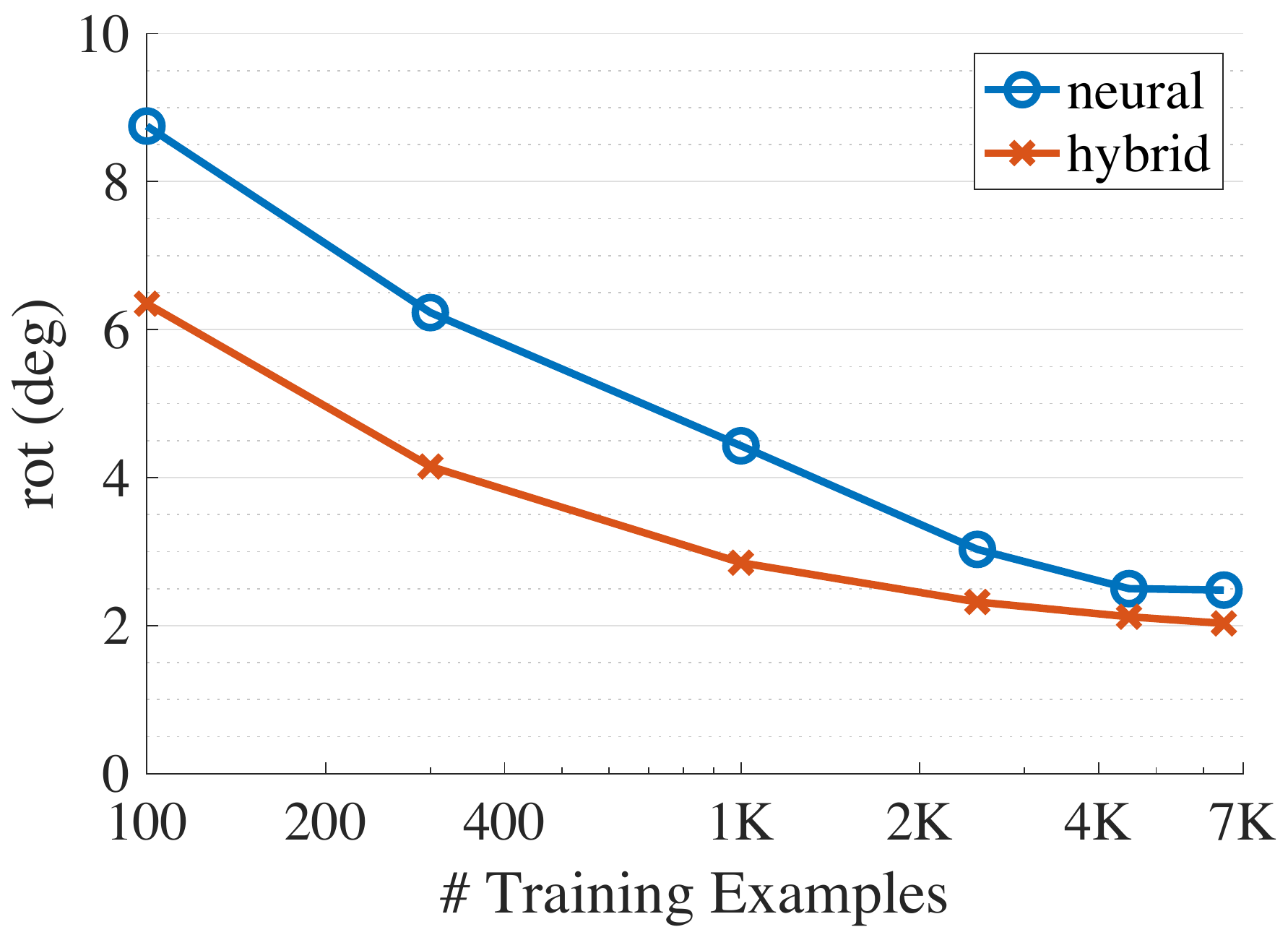}
    \caption{rotation (deg)}
    \end{subfigure}
    \vspace{-2pt}
    \caption{\small{Prediction errors \vs training data size. Our hybrid model not only performs better, but also requires much less data to achieve a given level of performance. In contrast, purely using purely data-driven models requires a larger training set and is not performing as well.}} 
    \label{fig:results_data}
    \vspace{-18pt}
\end{figure*}

\noindent{\bf Setup. }
For experiments on MIT push dataset, $\varphi^x$ and $\varphi^u$ consist of two bilinear layers with hidden sizes as 32 and 16 respectively and both followed by tanH activation. $\varphi^z$ is a single linear layer with hidden size 16 followed by tanH activation. $f^{\text{RNN}}_{\text{enc}}$ and $f^{\text{RNN}}_{\text{prior}}$ are both GRUs with 2 hidden layers and a hidden size of 16, followed by a linear layer of hidden size 16 for mean, and another parallel linear layer of hidden size 16 with softplus activation for standard deviation. $f^{\text{RNN}}_{\text{dec}}$ is a GRU with 2 hidden layers and a hidden size of 16, followed by a linear layer of hidden size 4 for mean, and another parallel linear layer of hidden size 4 with softplus activation for standard deviation. We again use ADAM optimizer~\cite{Kingma2015Adam} with a learning rate of $10^{-3}$ and a weight decay of 0.5 every 5,000 iteration for a total of 50,000 iteration. Our training set contains 6,500 trajectories, while our test set contains 628 trajectories.

\begin{table}[t]
    \centering
    \small
    \begin{tabular}{llccc}
    \toprule
    Materials & Models & trans (\%) & pos (mm) & rot (deg) \\
    \midrule
    \multirow{4}{*}{plywood} & Zero & 99.99 & 339.12 & 48.36 \\
    & Physics & 2.51 & 5.49 & 10.38 \\
    & Neural & 0.92 & 3.43 & 2.16 \\
    & Hybrid & 0.77 & 2.16 & 1.65 \\
    \midrule
    \multirow{4}{*}{delrin} & Zero & 99.99 & 357.98 & 52.67 \\
    & Physics & 1.89 & 5.78 & 12.07 \\
    & Neural & 0.81 & 2.81 & 2.50 \\
    & Hybrid & 0.62 & 2.09 & 2.19 \\
    \bottomrule
    \end{tabular}
    \caption{Our Hybrid model performs well consistently across object materials. Here for the rectangle made of plywood and delrin, our model again outperforms all other baseline models.}
    \vspace{-18pt}
    \label{tbl:results_push_multi}
\end{table}

\noindent{\bf Results. }
\tbl{tbl:results_push} shows the main results on the MIT Push data-set, using the \emph{rect1} object, a 837g square with a side length of 9cm, on the ABS surface. Our full model (Hybrid) significantly outperforms the baseline methods that rely only on physics engines or neural nets. Here we list results on both training and test sets. A pure neural net--based approach achieves a relatively low error on the training set, close to our Hybrid model. However, it generalizes much worse to the test set. Its prediction errors are much higher for both position estimation (2.81 \vs 2.04) and rotation estimation (2.48 \vs 2.03).

Our formulation is not constrained by the object's material. \tbl{tbl:results_push_multi} shows results on objects made of two other materials: plywood and delrin. Our Hybrid model consistently outperforms the baselines. 

Our hybrid model is also more sample-efficient. As shown in \fig{fig:results_data}, compared with the Neural model, our Hybrid model not only has higher prediction accuracies, but also achieves such accuracies much faster. Our model converges with as little as 2,500 training examples; in contrast, even with 6,500 training examples, purely data-driven models are not able to achieve performance comparable to ours.

\begin{figure}[t]
    \centering
    \includegraphics[width=\linewidth]{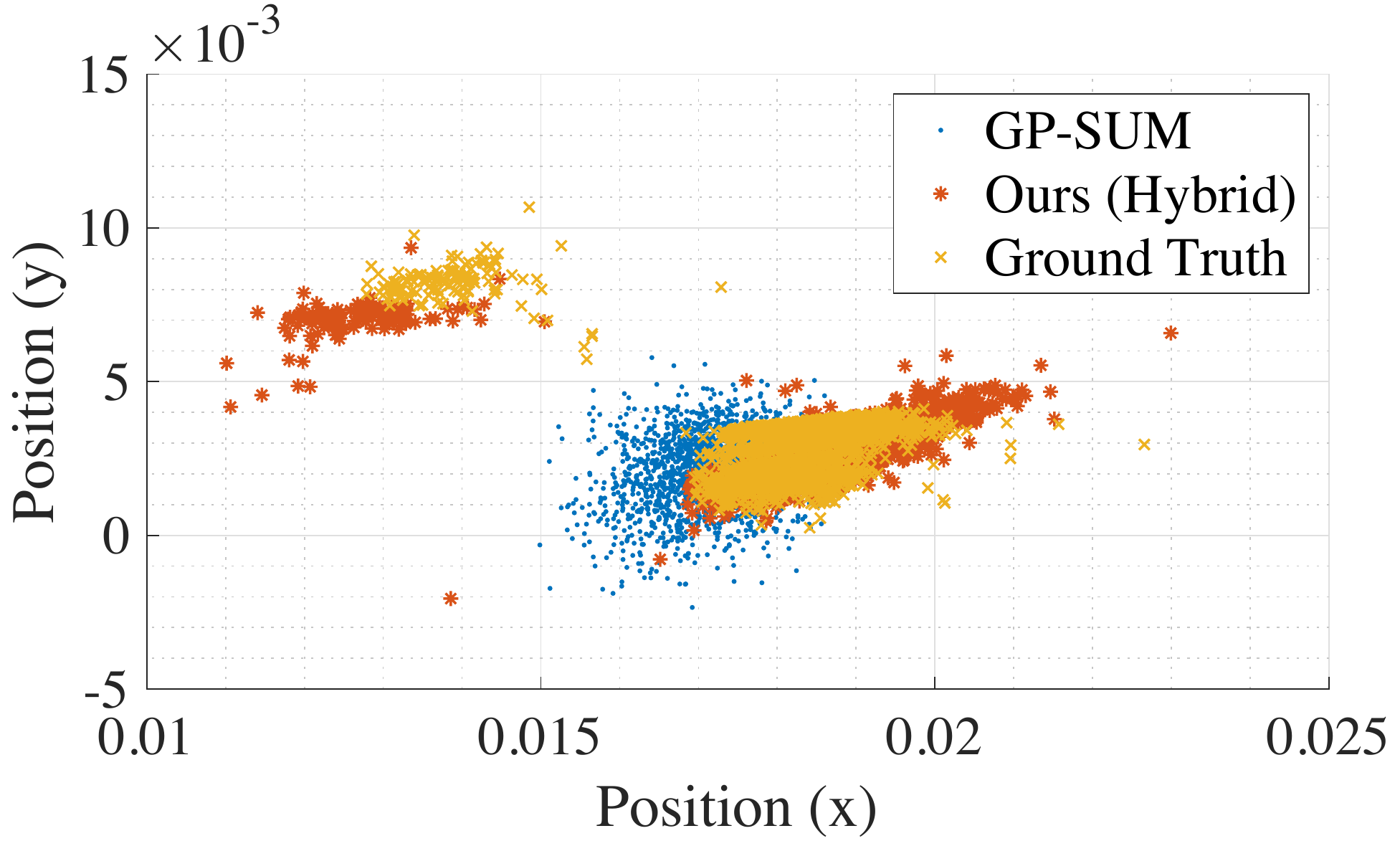}
    \vspace{-8pt}
    \ \\
    \small
    \begin{tabular}{lcc}
    \toprule
    & GP-SUM~\cite{bauza2017gp} & Ours (Hybrid) \\
    \midrule
    Chamfer Distance ($\times 10^{-4}$) & 6.80 & 2.77 \\
    \bottomrule
    \end{tabular}
    \caption{Our Hybrid model captures the distribution of possible push outcomes. Measured in Chamfer distance, our model achieves a lower error compared with GP-SUM~\cite{bauza2017gp}.}
    \vspace{-20pt}
    \label{fig:results_vis}
\end{figure}

Our model captures the uncertainty of the object motion well. To evaluate this, from the repeated pushes, we collect the ground truth distribution of the position of the object (\emph{rect1}) after being pushed for one second. We then sample 2,000 points from the state-of-the-art stochastic motion modeling approach---GP-SUM~\cite{bauza2017gp}. We also sample 2,000 trajectories from our Hybrid model. We present qualitative and quantitative results in \fig{fig:results_vis}. Compared with GP-SUM, our model can better capture the underlying uncertainty. Quantitatively, we compute the Chamfer distance~\cite{Barrow1977Parametric} between each model's output distribution $S$ and the ground truth distribution $T$, defined as
\begin{equation}
\small
    \text{CD}(S, T) = \frac{1}{|S|} \sum_{p \in S} \min_{q \in T} \norm{p - q}_2 + \frac{1}{|T|}\sum_{q \in T} \min_{p \in S} \norm{p - q}_2.
\end{equation}
Our model achieves a lower error compared to GP-SUM.

\subsection{Generalization Power}

We want our prediction models to generalize to real-world objects, which can be of any shape and material. In this section, we evaluate how our model and the baselines generalize to new object materials and shapes.

\begin{figure*}[t]
    \centering
    \begin{subfigure}[b]{0.37\linewidth}
    \includegraphics[width=\linewidth]{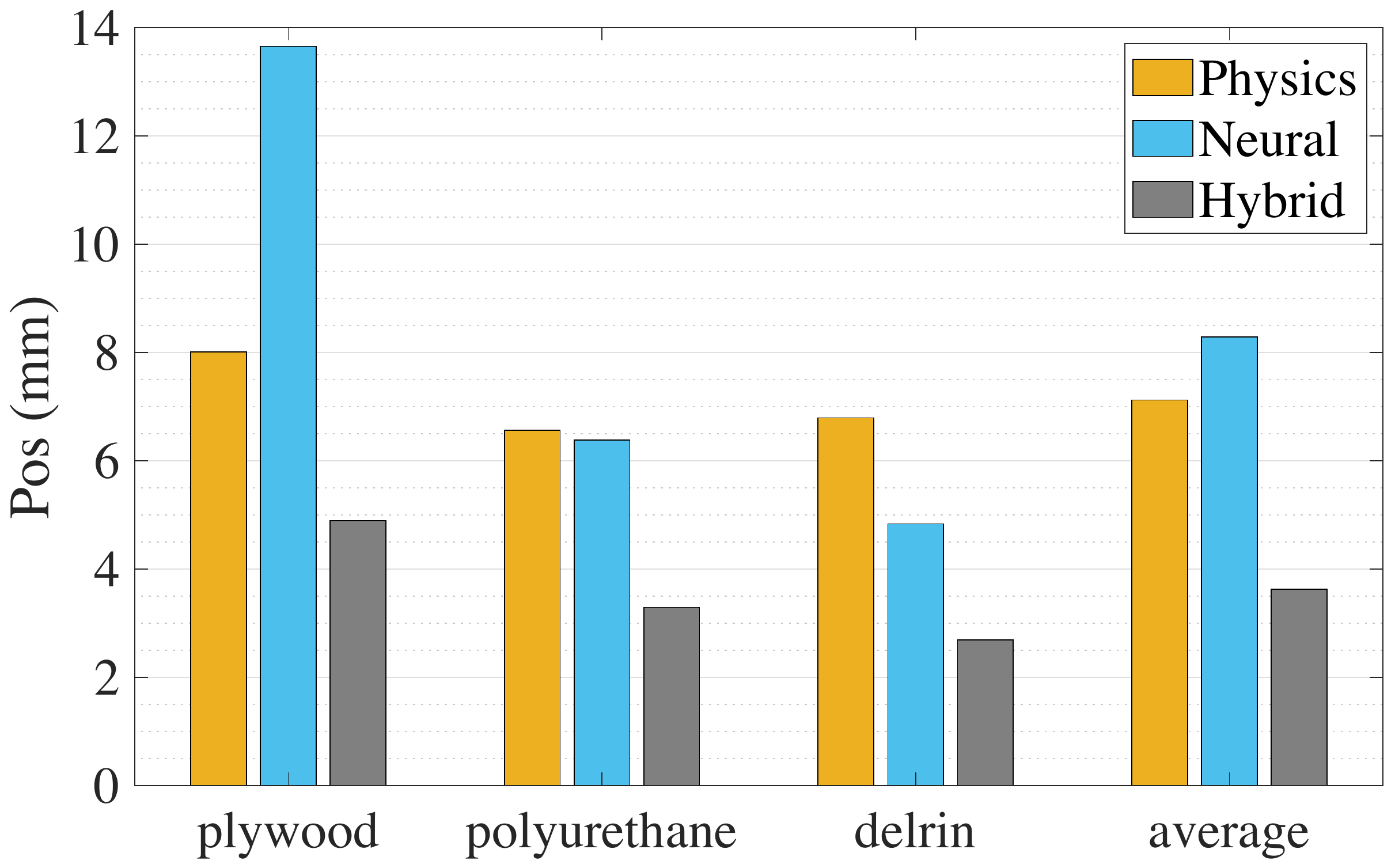}
    \caption{Position prediction on different materials}
    \label{fig:generalize_mat_1}
    \end{subfigure}
    \begin{subfigure}[b]{0.40\linewidth}
    \includegraphics[width=\linewidth]{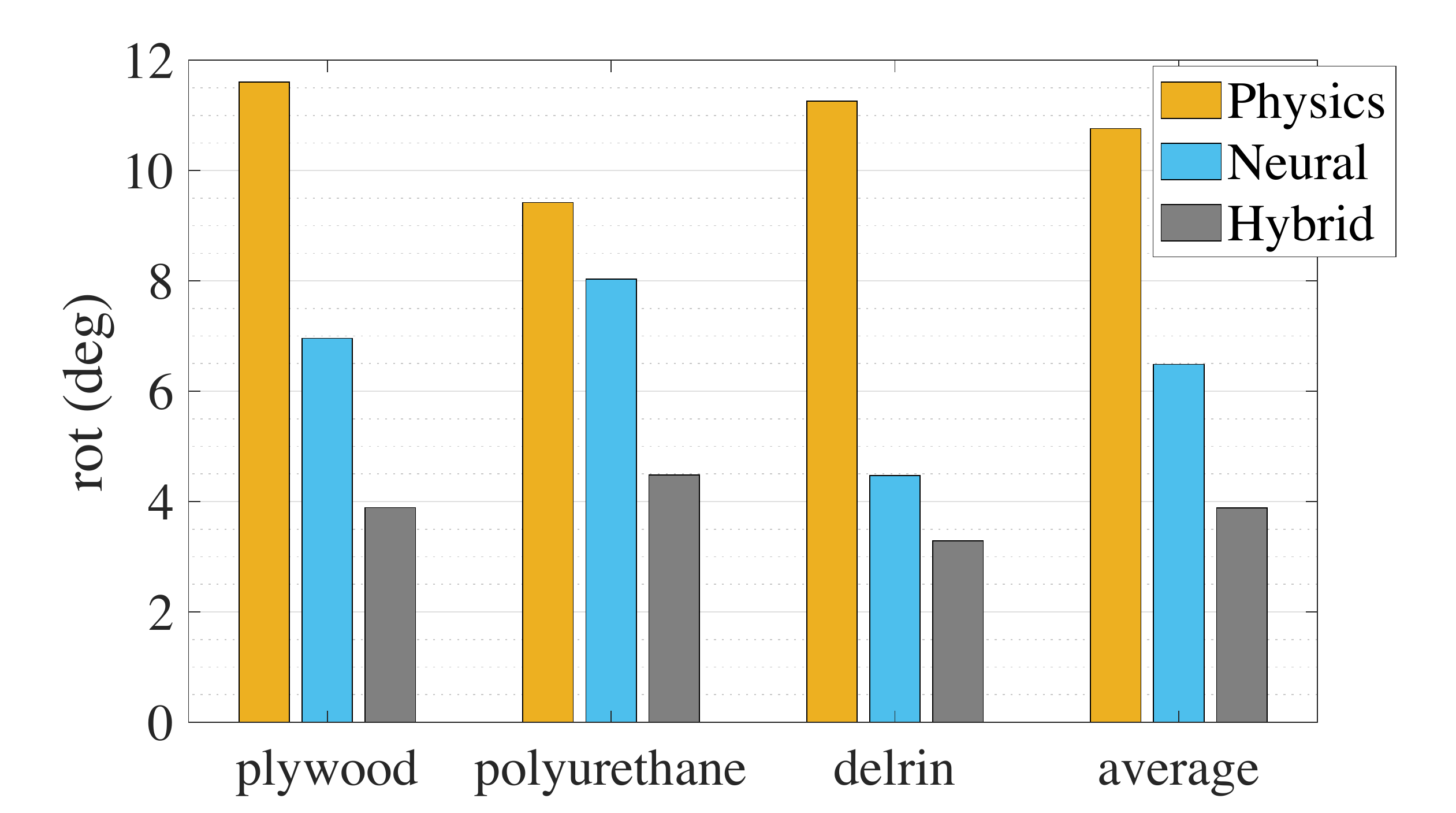}
    \caption{Rotation prediction on different materials}
    \label{fig:generalize_mat_2}
    \end{subfigure}
    \begin{subfigure}[b]{0.20\linewidth}
    \includegraphics[width=\linewidth]{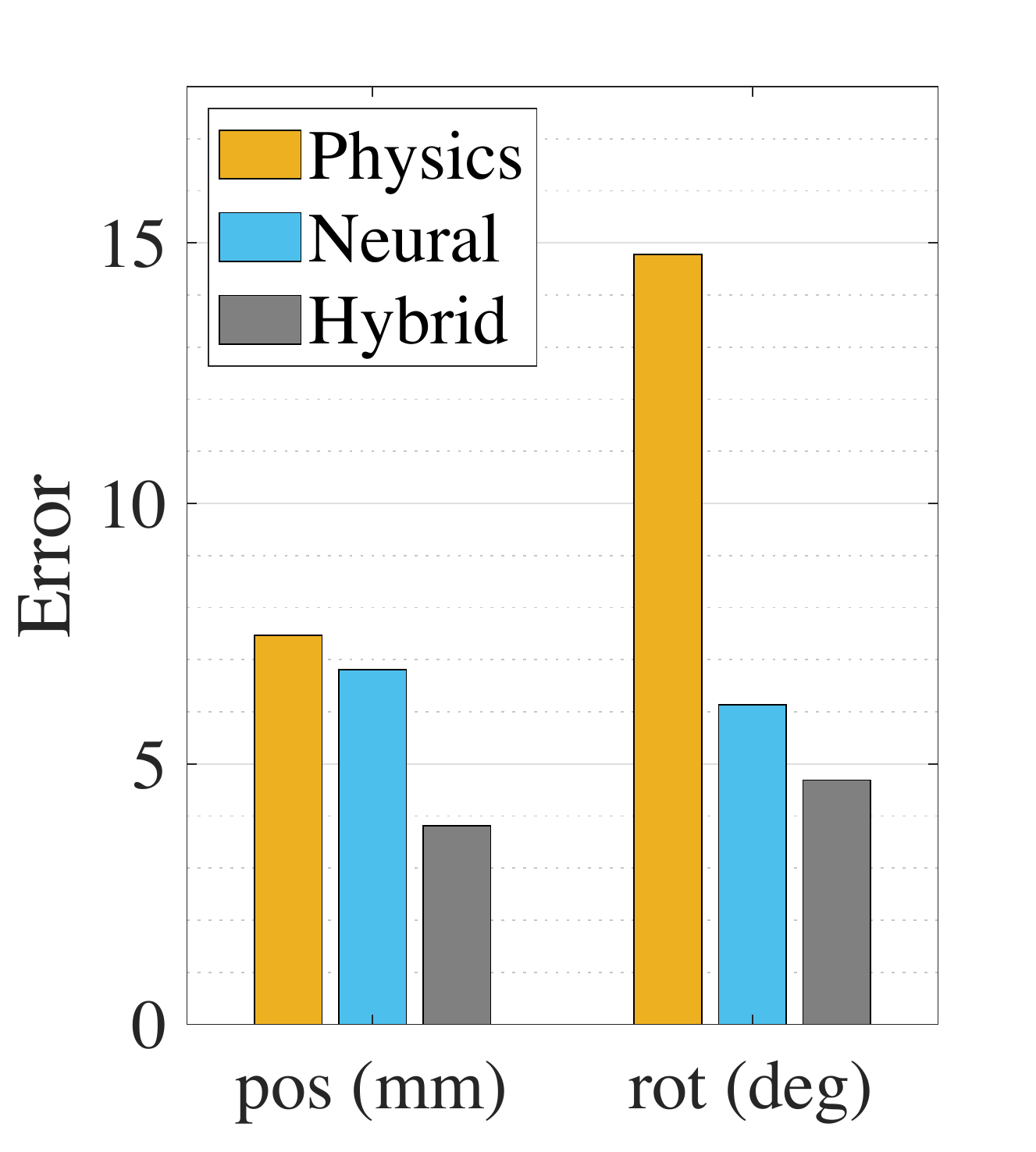}
    \caption{Shape generalization}
    \label{fig:generalize_shape}
    \end{subfigure}
    \vspace{-3pt}
    \caption{Our method generalizes well to different materials and to a new shape (\emph{rect2} in the push dataset). The error bars show our hybrid model achieves a consistently lower generalization error for both position and rotation prediction, compared with baseline methods.} 
    \vspace{-20pt}
\end{figure*}

For materials, we evaluate our models predictive abilities on different surfaces. We consider the push data-set on plywood, polyurethane, and delrin. \figs{fig:generalize_mat_1} and \ref{fig:generalize_mat_2} summarize the results of our Hybrid model and of the Physics and Neural baselines. Our model has lower generalization errors in both position and rotation prediction. 

For shapes, we evaluate our model's predictions on a new object---\emph{rect2}, a 1045g rectangle with side lengths of 9cm and 11.26cm. \fig{fig:generalize_shape} suggests that our model also generalizes better than the Physics and the Neural baselines on both position and rotation estimation.

We hypothesize that our model generalizes better across shapes and materials because it learns residuals---errors of the physics models that are supposed to be similar across various regimes. This assumption critically depends on the quality of the physics model; when it no longer holds, our model's generalization power may be limited.

\section{Discussion and Conclusion}
\label{sec:discussion}

We have proposed a simulation framework using a data-augmented residual motion model. Our underlying philosophy is to first exploit analytical models to model real-world data as much as possible, and then to learn the remaining residuals. This residual learning formulation adapts the model to specific real-world scenarios, with little need for domain specific knowledge or hand-crafting. In this study, we have demonstrated its efficacy in predicting real-world planar pushing and its generalization power across shapes and materials. The particular choice of our residual learner enables accurate long-term predictions and generates complex posterior distributions over future states. The improved accuracy may be attributed to the model's implicitly learning of the details in object-surface frictional interactions, so that it can account for variations in the coefficient of friction and potentially anisotropic friction.

One may be tempted to do away with the analytic model entirely and rely on a purely data-driven approach. This approach results in learning the full motion model from scratch with no priors. Aside from being more data-hungry, this approach does not generalize well to other shapes and materials. Starting from a good initial guess to the trajectory, the residual model has less to learn and generalizes better.

No model will ever be perfect; therefore, our model's ability in reasoning about possible outcomes can be a powerful tool in planning and control. For example, in the context of the planar pushing task, identifying a predictable push can lead the planner to exploit this property.

While the data-augmented residual model is more accurate and reasons about uncertainty, it does have certain drawbacks. Given that the residual model is data-driven, it can predict outcomes that are physically impossible, because there is no mechanism to enforce basic physics principles (\eg, the conservation of energy) at the output level. In practice, however, these implausible outcomes would not appear if we can have the model sufficiently trained.

The simulation framework only requires a coarse domain-specific analytical motion model to be applied to other robotic tasks. For future work, we plan on applying the framework to more difficult modeling tasks. In this paper's experimental setup, we have assumed access to the noisy full state space; an interesting extension would be to learn an observation model along with the residual model to simultaneously perform state estimation and prediction.

\addtolength{\textheight}{-0cm}   


\noindent\textit{Acknowledgement } This work is supported by NSF \#1420316, \#1523767, and \#1723381, AFOSR grant FA9550-17-1-0165, ONR MURI N00014-16-1-2007, Toyota Research Institute, Honda Research, Facebook, and Draper Laboratory. Any opinions, findings, and conclusions or recommendations expressed in this material are those of the authors and do not necessarily reflect the views of our sponsors.

\bibliographystyle{IEEEtran}
\bibliography{physplus_iros}

\end{document}